\documentclass{article}

\usepackage[preprint]{neurips_2025}
\usepackage{diagbox} % 关键宏包：提供斜线分割命令
\usepackage{multirow} % 支持多行合并

\usepackage[utf8]{inputenc} % allow utf-8 input
\usepackage[T1]{fontenc}    % use 8-bit T1 fonts
\usepackage{hyperref}       % hyperlinks
\usepackage{url}            % simple URL typesetting
\usepackage{booktabs}       % professional-quality tables
\usepackage{amsfonts}       % blackboard math symbols
\usepackage{nicefrac}       % compact symbols for 1/2, etc.
\usepackage{microtype}      % microtypography
\usepackage{xcolor}         % colors
\usepackage{graphicx}
\usepackage {amsmath}
\usepackage{amssymb}

\workshoptitle{Efficient Reasoning}

\title{Adaptive Dual Reasoner: Large Reasoning Models Can Think Efficiently by Hybrid Reasoning}

\author{
  Yujian Zhang$^{1,*}$ \\
  Tencent Youtu Lab \\
  \texttt{23s003081@stu.hit.edu.cn} \\
  \And
  Keyu Chen$^{1}$ \\
  Tencent Youtu Lab \\
  \texttt{yolochen@tencent.com} \\
  \AND
  Zhifeng Shen \\
  Tencent Youtu Lab \\
  \texttt{billsshen@tencent.com} \\
  \And
  Ruizhi Qiao \\
  Tencent Youtu Lab \\
  \texttt{ruizhiqiao@tencent.com} \\
  \And
  Xing Sun \\
  Tencent Youtu Lab \\
  \texttt{winfredsun@tencent.com} \\
}

\begin{document}

\maketitle

\vspace{-1em}
{\renewcommand{\thefootnote}{\arabic{footnote}}
\footnotetext[1]{Equal contribution.}
\renewcommand{\thefootnote}{\fnsymbol{footnote}}
\footnotetext[1]{Work done during internship at Tencent.}
}

\begin{abstract}
    % 尽管推理模型在各种推理任务取得比非推理模型更好的性能，但是推理模型却因为过度和冗长的思考过程导致计算量和推理时间的增加。主要原因尤其体现在推理模型思考步骤中存在着对于简单步骤进行过度思考，这导致模型在CoT的步骤的资源分配不均的问题。尽管一些现有工作通过设置prompt或者训练的方式让模型在全局思考过程进行长度精简，但这也引发对于需要深度思考的步骤分配资源不足，从而导致性能退化。为此，我们通过为推理模型设计了一种混合思考方式，支持两种思考模式：快思考和慢思考。快思考使用更精练的思考过程来让模型加速简单思考步骤的思考速度，而慢思考则保留原始推理模型深度思考能力，来全力解决困难的思考步骤。这种交叉思考模式可以使得模型根据上下文的逻辑复杂性自适应地调节思考深度，进行合理思考资源分配。我们方法包括两个步骤，首先我们构建了一种混合思考数据构建流水线用于模型第一阶段的冷启动微调，让模型过度到混合思考模式。在第二阶段，我们通过强化学习方法来进一步优化混合思考的能力，我们设计了一种基于熵引导的动态roll out方式，该方式鼓励模型在遇到熵值较高的慢思考节点进行分支，来鼓励模型积极的探索。此外，我们还针对问题难度来对探索路径中快思考和慢思考步骤进行动态地惩罚，用于促进模型在思考资源的合理分配。广泛的实验结果表明，我们方法在推理任务具有更快和更好效果，优于现有的方法，...
    
Although Long Reasoning Models (LRMs) have achieved superior performance on various reasoning scenarios, they often suffer from increased computational costs and inference latency caused by overthinking. %This inefficiency is primarily caused by the model’s tendency to over-elaborate on reasoning steps, leading to uneven resource allocation across the different steps of the reasoning process. %While existing approaches attempt to shorten the overall reasoning length via prompt engineering or post-training techniques, such global refinements frequently result in inadequate allocation of computational resources for more challenging steps, ultimately impairing model performance.
To address these limitations, we propose an \textbf{A}daptive \textbf{D}ual \textbf{R}easoner, which supports two reasoning modes: fast thinking and slow thinking. \textbf{ADR} dynamically alternates between these modes based on the contextual complexity during reasoning. %, enabling efficient allocation of computational resources.
%Fast thinking enables the model to efficiently handle straightforward steps with concise reasoning, while slow thinking retains the model’s original deep reasoning capability to tackle complex reasoning steps. This alternating paradigm allows the model to dynamically adjust its reasoning depth according to the logical complexity of the context, thereby achieving optimal allocation of computational resources.
\textbf{ADR} is trained in two stages: 
% (1) A hybrid reasoning data construction pipeline provides high-quantity training data for the cold-start stage, which enables the model to learn and integrate both modes of reasoning; (2) For the reinforcement learning stage, we employ an entropy-guided dynamic rollout strategy, which facilitates the model to branch at high-entropy units for deeper exploration. Additionally, a difficulty-aware penalty mechanism is exploited to regulate the proportion of fast and slow thinking steps, encouraging rational allocation of reasoning resources.
(1) A cold-start stage using supervised fine-tuning (SFT) to equip the model with the ability to integrate both fast and slow reasoning modes, in which we construct a hybrid reasoning dataset through a dedicated pipeline to provide large-scale supervision. (2) A reinforcement learning stage for optimizing reasoning effort, where we introduce \textbf{E}ntropy-guided \textbf{H}ybrid \textbf{P}olicy \textbf{O}ptimization (\textbf{EHPO}), an RL training framework employing an entropy-guided dynamic rollout strategy for branching at high-entropy units and a difficulty-aware penalty to balance fast and slow reasoning.
Across challenging mathematical reasoning benchmarks, \textbf{ADR} achieves an effective balance between reasoning performance and efficiency among state-of-the-art approaches. Specifically, \textbf{ADR} yields a performance gain of up to 6.1\%, while reducing the reasoning output length by 49.5\% to 59.3\%.

  % The abstract paragraph should be indented \nicefrac{1}{2}~inch (3~picas) on
  % both the left- and right-hand margins. Use 10~point type, with a vertical
  % spacing (leading) of 11~points.  The word \textbf{Abstract} must be centered,
  % bold, and in point size 12. Two line spaces precede the abstract. The abstract
  % must be limited to one paragraph.
\end{abstract}

\section{Introduction}

% 随着近年来long reasoning models (LRMs)的出现[ref deepseek, qwen3-think， gpt5]，让深度思考这一推理范式在各种高难度的数学推理，逻辑推理等任务成为主流解决方案，并展现优异的性能表现。尽管这种深度chain-of-thought 思考能够让模型实现推理进行深度或者广度的探索，但是由于大模型本身还是基于next-token-predection的串行解码模式，这对于需要输出大量思考token的LRMs无疑会增大计算资源和时间延迟。尽管有类似推测解码[ref 1,2]、并行解码工作[ref 3,4]致力于提升模型解码速度，但是还是无法从根本上解决模型思考过程过长的问题。此外，一个问题是，LRMs经常出现over-think，通常表现在对于一些简单子问题解决使用过度思考资源，这就导致LRMs在面对不同难度的子问题的解题步骤中出现思考资源分配不平衡的问题,这种思考资源的过度挥霍加剧了计算资源的消耗。[ref 1,2,3]

With the recent emergence of Long Reasoning Models (LRMs) \citep{Deepseek-r1,Seedthinking,Qwen3}, the Chain-of-Thought (CoT) \citep{Chain-of-thought} reasoning has been further popularized as the mainstream paradigm for tackling complex tasks such as mathematical or logical problems. %, demonstrating remarkable performance.
However, LRMs are notorious for over-thinking \citep{StopOverthinking, su2025between}, wherein the model unnecessarily generates redundant reasoning.
%While long CoT reasoning enables models to explore reasoning trajectories in both depth and breadth, LRMs are still based on the next-token prediction paradigm during inference. This serial process imposes considerable computational overhead and latency when generating extremely long reasoning sequences. Although recent advancements, such as speculative decoding \citep{leviathan2023fast, cai2023medusa, li2025eagle} and parallel decoding methods \citep{chen2025aspd, yang2025multiverse}, have been proposed to accelerate the decoding speed, these approaches do not aim to address the issue of the excessive length of reasoning sequences. Furthermore, LRMs are notorious for over-thinking \citep{StopOverthinking, su2025between}, wherein the model unnecessarily generates redundant reasoning. %This leads to imbalanced distribution of computational effort across sub-tasks of varying complexity, ultimately aggravating the overall consumption of computational resources [ref 1,2,3].
% 最近涌现了许多的工作专注于解决LRMs过度思考的问题[ref]，一种最为简单的方式是基于提示工程，它们通过特定让模型输出步骤内容保持精简来缩短整体的思考长度[ref cod, concise], 虽然输出长度得到较大幅度的缩短，但是由于这种全局作用方法使得需要深度思考的复杂步骤产生under-thinking问题，导致解决难度较大问题时候性能下降。一些基于token采样概率操纵的方法[ref 5,6]，通过降低特定思考切换词（例如Wait，Hmm）的概率避免模型思维切换，但是存在跳过那些深度思考关键步骤的风险。基于token预算方法[ref 7,8]通过让模型在特定token预算下进行思考长度控制的表现来提高推理效率,还有些方法通过计算特定信号值让模型提前退出思考来提前得到答案[ref 11,12]，或者让模型根据难度模型自动判断选择是否进行显示深度思考或直接产生解决方案，这些方法仍在思考过程的粒度能需要比较精确控制。而基于强化学习的促进LRMs高效推理方式受到广泛的关注[ref 9,10]，。
Recently, a surge of research has focused on addressing the overthinking problem in LRMs. One of the simplest approaches \citep{CoD,CCoT,TokenComplexity} is prompt engineering, which aims to make the model’s output steps more concise through specific prompts. 
Other methods, such as probability manipulation \citep{TIP}, token budget \citep{Token-budget,TokenComplexity,Selfbudgeter, muennighoff2025s1}, early exiting \citep{yang2025dynamic,fu2024efficiently,zhao2025let}, and CoT compression \citep{kang2025c3ot,xia2025tokenskip}, focus on avoiding frequent shifts in thought or shorten the output length. However, these length-driven approaches may lead to insufficient exploration of complex reasoning steps that require deeper thinking. %, ultimately resulting in degraded performance on more challenging problems.
% 为了进一步进一步控制LRMs对应不同问题的推理长度，许多基于强化学习的方法被提出来用以缩短和精练思考过程\cite{REO-RL，L1-exact，ShorterBetter，ConciseRL，ConciseReasoning,LASER-D,Self-Adaptive,LC-R1,Bingo,MinD}。ShorterBetter \citep{ShorterBetter} 让模型通过自我监督的方式动态地发现每个问题的最优推理长度，从而在保持准确性的前提下，显著减少不必要的推理步骤。L1 \citep{L1-exact} 提出了一种Length Controlled Policy Optimization 使得模型能够在不同的计算预算下灵活调整推理链的长度，但在一些跨领域（OOD）任务中，长度控制的误差仍然较高。ConciseRL \citep{ConciseRL} 通过一个大型语言模型作为评估器（judge）来动态评估推理痕迹的简洁性，从而避免了传统基于token数量的奖励方法的局限性。LASER-D \citep{LASER-D} 使用一个基于目标长度的阶跃函数作为奖励，旨在通过奖励机制鼓励模型生成更短但正确的推理链。Self-Adaptive \citep{Self-Adaptive}算法基于样本对之间的比较来构建奖励，而不是直接对回答长度进行惩罚。LC-R1 \citep{LC-R1}使用一个轻量级的解析器 LC-Extractor，从原始推理过程中提取有效部分，生成压缩后的推理序列，通过计算压缩后序列的相对Length Reward让推理长度匹配自动调整到问题的难度，并结合Compress Reward鼓励模型在得出正确答案后立即停止推理。BINGO \citep{Bingo}使用重要性的长度奖励旨在区分重要和不重要的token，并仅对不重要的token进行惩罚，从而在不降低性能的情况下减少冗余，并结合动态长度奖励在训练过程中调整奖励信号来平衡模型在解空间探索和压缩的需求。
To further refine the control of reasoning length in LRMs for different problems, a variety of reinforcement learning methods have been proposed \citep{L1-exact, ShorterBetter, ConciseRL, Self-Adaptive, LC-R1} to perform various length or difficulty-based rewards. % ShorterBetter \citep{ShorterBetter} guides the model to dynamically discover the optimal reasoning length for each problem with the Sample Optimal Length metric. 
To control reasoning behaviors, a range of approaches \citep{Thinkless,wu2025arm,tu2025learning,Adaptthink,jiang2025think,zhang2025continue,lou2025adacot} has introduced the concept of hybrid reasoning modes (\textit{e.g.}, fast thinking, and slow thinking). %, which aim to improve inference efficiency. 
Some hybrid reasoning methods utilize a router to select appropriate models \citep{gpt5} or reasoning modes \citep{aytes2025sketch,liang2025thinkswitcher} according to the estimated difficulty of the query. Subsequent works have sought to eliminate the dependency on routers by adopting reinforcement learning frameworks, enabling autonomously routing to select the appropriate reasoning mode \citep{jiang2025think,zhang2025continue,lou2025adacot}. Nevertheless, actual reasoning trajectories often comprise sub-problems of varying complexity, and coarse-grained control over reasoning modes is unable to adaptively allocate cognitive resources for each reasoning path. %, resulting in inefficient distribution of reasoning effort. 
To tackle this issue, recent approaches \citep{MinD,DTO,ACPO} decompose reasoning steps into smaller units for fine-grained control, but they rely on static rollout strategies that restrict deeper exploration on hard subproblems.To overcome this limitation, we propose the \textbf{A}daptive \textbf{D}ual \textbf{R}easoner, which dynamically switches between fast and slow reasoning modes according to contextual complexity. \textbf{ADR} is trained in two stages: first, a supervised fine-tuning stage equips the model with both reasoning modes; second, we introduce \textbf{E}ntropy-guided \textbf{H}ybrid \textbf{P}olicy \textbf{O}ptimization (\textbf{EHPO}), a reinforcement learning framework that leverages entropy trends for dynamic rollout strategy and a difficulty-aware penalty to balance efficiency and accuracy. Our contributions are summarized as follows:
\begin{itemize}

\item \textbf{ADR}, a novel reasoning paradigm, which enables large reasoning models to adaptively switch between fast reasoning for straightforward cases and slow reasoning for complex dependencies, laying the foundation for flexible allocation of reasoning effort.

\item An automated hybrid reasoning data construction curator. We build a scalable pipeline that automatically constructs hybrid reasoning data, enabling existing LRMs to transition smoothly into the new hybrid reasoning paradigm.

\item \textbf{EHPO}, a reinforcement learning framework that integrates entropy trends for dynamic rollout strategy and a difficulty-aware penalty to balance efficiency and accuracy, thereby optimizing reasoning under the hybrid paradigm.

\end{itemize}

%Although this method considerably reduces output length, its global control nature can cause under-thinking during complex steps that actually require deep reasoning, leading to degraded performance on more challenging problems. Some methods , based on manipulating token sampling probabilities, reduce the probability of certain reasoning transition words (e.g., “Wait”, “Hmm”) to avoid frequent shifts in thought. However, this comes with the risk of skipping critical steps necessary for deep reasoning. Token budget–based methods  control the length of the model's reasoning process by constraining it within a specific token budget to improve inference efficiency. Other approaches compute certain metrics to enable the model to terminate reasoning early to output an answer in advance , or allow the model to automatically decide whether to perform explicit deep reasoning or directly generate a solution based on the assessed problem difficulty. However, these methods still require relatively precise control over the granularity of the reasoning process. Recently, reinforcement learning–based approaches to encourage more efficient inference in LRMs have attracted significant attention[ref 9,10].

% Introduction Here.

% \section{Related Work}
% % 感觉都可以写在introducutin，剩下篇幅
% Related Work Here.

\section{Methodology}
% 本文提出了一种两阶段训练流程，使大型推理模型（LRM）能够在推理过程中根据上下文的逻辑复杂性自适应地调节思考深度：在逻辑推理较为直接时保持高效推理快速得出结论，而在逻辑关系更为复杂时开始深度思考，保证结论正确。具体而言，第一阶段为冷启动阶段，通过有监督微调（SFT）将LRM从原有的“<think>t</think>a”模式转换为动态切换快慢思的考模式，使其初步具备根据上下文逻辑复杂性选择合适的思考模式的能力；第二阶段为强化学习（RL）阶段，通过引入基于难度的思考模式奖励机制和基于熵值的动态采样策略，优化模型的资源分配策略，在保持模型原有推理性能的前提下，提升模型的推理效率。
% We propose a two-stage training framework enabling large reasoning models (LRMs) to adaptively allocate reasoning effort based on contextual complexity: employing fast reasoning for straightforward cases and intensive reasoning for complex dependencies to ensure correctness. In the first stage (cold-start), supervised fine-tuning (SFT) guides the model to distinguish between “fast” and “slow” reasoning modes and to select the appropriate mode. In the second stage (reinforcement learning), a difficulty-aware reward combined with entropy-based dynamic sampling optimizes resource allocation, promoting fast reasoning where suitable without compromising accuracy, thus enhancing overall efficiency.
\begin{figure}[htbp]
    \centering
    \includegraphics[width=0.7\textwidth]{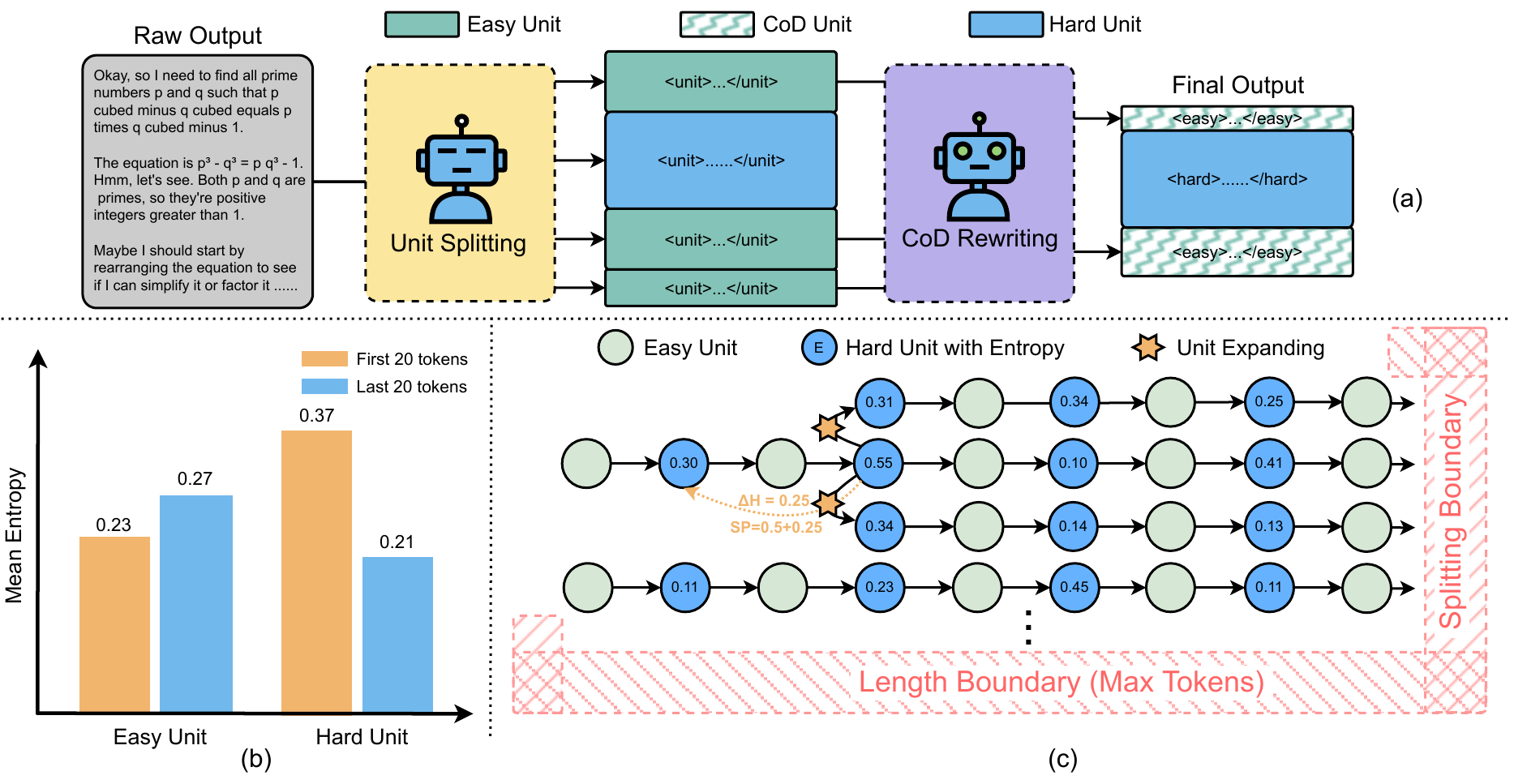}
    \caption{(a) Hybrid Reasoning Data Construction. (b) Entropy Analysis of Two Reasoning Units: Transitions from easy to hard mode exhibit higher entropy. (c) Entropy-Guided Dynamic Rollout Strategy: Branching occurs with probability $\text{SP} = \alpha + \Delta H$ when transitioning from easy mode to hard mode, where $\Delta H$ denotes the normalized entropy difference.}
    \label{fig:ourmethod}
\end{figure}
\subsection{Aligning the Model to the Adaptive Dual Reasoning Paradigm}

% Previous work \citep{wang2025beyond, lin2025controlling} shows that keywords related to reflection, verification, and exploration correlate with higher entropy, indicating greater uncertainty and the demand for reasoning effort. Inspired by this, we decompose trajectories into reasoning units, labeling those with such content as hard and others as easy. Easy units are compressed with CoD-style reasoning to reduce token usage, while hard units remain uncompressed. The two categories are annotated with special tokens to form the final reasoning format. The complete data construction process is illustrated in Figure \ref{fig:ourmethod}(a).

To align the model with the adaptive dual reasoning paradigm that supports both fast and slow reasoning modes, we conduct cold-start training via supervised fine-tuning (SFT) and construct hybrid reasoning data from an open-source reasoning dataset. Inspired by the observation that higher entropy is associated with keywords related to reflection, verification, and exploration\citep{wang2025beyond, lin2025controlling}, we propose a data construction process based on CoT decomposition and rewriting, as shown in Figure \ref{fig:ourmethod}(a). Specifically, we decompose reasoning trajectories into reasoning units, labeling those with high-entropy content as hard and others as easy. Easy units are compressed using CoD-style to minimize token usage, while hard units remain uncompressed to retain reasoning depth. These units are then annotated with special tokens to form the final reasoning format as following:

\begin{equation}\label{eq1}
\text{<think> }\text{<easy> }u_{1}\text{ </easy> }\text{<hard> }u_{2}\text { </hard> } \cdots \text { <easy> }u_{n}\text{ </easy>}\text{ </think> }a
\end{equation}

\subsection{\textbf{E}ntropy-Guided \textbf{H}ybrid \textbf{P}olicy \textbf{O}ptimization}

To further improve reasoning efficiency while preserving accuracy, we propose \textbf{EHPO}, a reinforcement learning framework that updates the model using a GRPO-based objective. EHPO combines mode control reward with entropy-guided dynamic rollout to suppress unnecessary deep reasoning while retaining essential hard units across problems of varying difficulty.

% Interleaved Reasoning Policy Optimization - IRPO 有了 Iterative Reasoning Preference Optimization
% Entropy-guided Hybrid/Dynamic Policy Optimization - EHPO/EDPO

% 经过冷启动训练后，模型已经能够清楚区分快慢两种推理模式，并在推理过程中动态切换思考模式。但不同于CoD的思考模式下模型会大量减少反思、验证和探索相关的内容，SFT阶段仅仅对CoT中无需深度思考的部分进行简写，但没有精准去除原始CoT中冗余的深度思考思考单元。因此，在对其混合思考模式后，我们利用RL优化模型的推理资源分配，使其减少无用的深度思考单元以提升推理效率

% After cold-start training, the model can differentiate between shallow and intensive reasoning modes and dynamically switch between them. However, the SFT stage only simplifies CoT segments that do not require intensive reasoning, leaving redundant intensive units largely intact. Building on this, we employ RL to further optimize the allocation of reasoning effort, reducing unnecessary intensive reasoning while preserving essential units, thereby improving overall reasoning efficiency.
% After cold-start training, we further adopt RL training to refine reasoning allocation, suppressing unnecessary intensive reasoning while retaining essential units to improve efficiency.

\subsubsection{Reward Design}

% 我们基于GRPO，设置了由四种奖励信号组成的奖励函数，来针对模型的推理资源分配进行联合优化
% Building upon GRPO, we design a reward function comprising four distinct reward signals, which jointly optimize the model’s allocation of reasoning resources:
% To further reduce redundant intensive reasoning while preserving essential steps, we employ RL training to optimize reasoning efficiency. Building upon GRPO,
We design a reward function with four signals to jointly optimize the allocation of reasoning effort:

\begin{equation}\label{eq2}
    R=\mathcal{R}_{\text {format }}*\mathcal{R}_{\text {accuracy }}*\mathcal{R}_{\text {unit }}*\mathcal{R}_{\text {mode }}
\end{equation}

% 其中，除了用于监督模型回复正确性的\mathcal{R}_\mathrm{accuracy}和用于监督模型输出格式的\mathcal{R}_\mathrm{format}外，我们额外设置了1）单元语义奖励和2）模式控制奖励来监督优化模型合理使用两种思考模式。
% In addition to the accuracy reward $\mathcal{R}_\mathrm{accuracy}$ for supervising response correctness and the format reward $\mathcal{R}_\mathrm{format}$ for enforcing structural compliance, we further define (1) a unit-level semantic reward and (2) a mode-control reward to regulate the appropriate utilization of the two reasoning modes.

While the first two rewards enforce structural compliance and correctness, we highlight the latter two below.

% 单元语义奖励用于监督模型对两种思考模式的区分度，防止模型在训练过程中混淆两种思考模式，退化为原始的思考模式。我们对模型回复中的每个单元进行关键词检测，按照以下规则标记每个单元的语义正确性：1)模型标记为easy的单元，不允许出现表示反思、验证和探索的关键词；2)模型标记为hard的单元，必须包含表示表示反思、验证和探索的关键词。
\paragraph{Unit semantic Reward} To encourage the model to distinguish between two modes of reasoning rather than collapsing into the original paradigm, we define a unit semantic reward based on keyword matching. 
% Each unit $u_i$ is assigned semantic correctness as:
Each reasoning unit $u_i$ is semantically correct only if (i) $u_i$ is easy and contains no reflection/verification keywords such as "Wait", "However" and "Alternatively", or (ii) $u_i$ is hard and contains at least one such keyword. Then, the overall unit semantic reward is defined as follows:

% \begin{equation}
%     C\left(u_{i}\right)=\left\{\begin{array}{ll}1, & \text { if } \operatorname{mode}\left(u_{i}\right)=\text { easy and } u_{i} \cap \mathcal{K}=\varnothing \\1, & \text { if } \operatorname{mode}\left(u_{i}\right)=\text { hard and } u_{i} \cap \mathcal{K} \neq \varnothing \\0, & \text { otherwise, }\end{array}\right.
% \end{equation}

% 其中\left\{u_{1}, u_{2}, \ldots, u_{n}\right\}为回复中的思考单元，\operatorname{mode}\left(u_{i}\right)\in \left\{easy,hard\right\}表示单元u_{i}的类型，\mathcal{K}为表示反思、验证和探索的关键词。只有当回复中所有单元均满足语义正确时，该回复的单元语义奖励为1，否则为0
% where $\operatorname{mode}(u_i)\in\{\text{easy},\text{hard}\}$ denotes the reasoning type of unit $u_i$, and $\mathcal{K}$ is the set of keywords representing reflection, verification, and exploration. A response receives unit reward only if all units are semantically correct:

\begin{equation}
    \mathcal{R}_{\text{unit}}=\left\{\begin{array}{ll}1, & \text { if all units are semantic correct,}  \\0, & \text { otherwise }\end{array}\right.
\end{equation}

% 为优化模型的token资源分配策略，我们设置模式控制奖励鼓励模型在简单任务上优先使用easy模式。得益于GRPO的实现机制，我们可以通过一个题目下一组回复的平均正确率估计改该题目的难度，并实现基于难度的模式控制奖励：
\paragraph{Mode Control Reward} 
% To optimize the model’s utilization of reasoning effort, we introduce a mode control reward designed to encourage the preferential use of the easy reasoning mode for tasks of lower difficulty. Leveraging GRPO’s inherent group-sampling mechanism, the difficulty of a given task can be estimated from the average correctness across a set of model responses. Furthermore, we formulate a difficulty-aware mode control reward as follows:
To optimize the model’s utilization of reasoning effort, we introduce a difficulty-aware mode control reward, which encourages the preferential use of the easy mode on lower-difficulty tasks while promoting deeper reasoning in the hard mode for challenging ones:

\begin{equation}
\mathcal{R}_{\text {mode }} = \beta + (1 - \beta) \cdot \left( \frac{N_{pass}}{N} \cdot p_{\text{easy}} + (1 - \frac{N_{pass}}{N}) \cdot p_{\text{hard}} \right),
\end{equation}

% 其中N和N_{pass}分别表示一个题目下采样总数和其中答案正确回复数量，p_{\text{easy}}和p_{\text{hard}}分别表示模型使用easy和hard模式生成token占全部CoT中token总数的比例。\beta用于控制模式控制奖励的权重，将其取值范围限制在[\beta, 1]范围内。
% where $N$ and $N_{\text{pass}}$ denote the total number of sampled responses for a given problem and the number of correct responses, respectively. $p_{\text{easy}}$ and $p_{\text{hard}}$ indicate the fractions of tokens in the CoT that are generated under the easy and hard reasoning modes, respectively. $\beta$ is a hyperparameter that determines the weight of the mode control reward, ensuring that the reward remains within the interval $[\beta, 1]$.
where $N$ and $N_\text{pass}$ denote total and correct samples, $p_\text{easy}$ and $p_\text{hard}$ are the token ratios of the easy mode and hard mode, respectively, and $\beta$ is a hyperparameter controlling the reward scale within the range  $[\beta,1]$. We set $\beta=0.7$ by default.

% 注意，我们并没有设置显式的奖励限制模型的输出长度，而是利用\mathcal{R}_{\text {mode }}鼓励模型快速思考来隐式地缩短模型的输出长度。
% Note that we do not impose an explicit constraint on output length. Instead, the mode-control reward $\mathcal{R}_{\text{mode}}$ implicitly encourages the model to favor the easy reasoning mode, thereby shortening outputs.This observation is further supported by the results presented in Section 3.2.

\subsubsection{Entropy-Guided Dynamic Rollout Strategy}

% 在使用上述奖励函数进行GRPO训练过程中，我们观察到抑制模型进行深度思考的\mathcal{R}_{\text {mode }}会干扰模型原本的探索行为，压缩模型的探索空间，影响模型回复的正确率。通过统计模型输出时的熵值变化，我们发现模型进行推理时存在如下规律：模型在从easy模式转换为hard模式时熵值往往高于由hard模式转换为easy模式时的熵值。而这也符合动态混合思考的设计原则：模型遇到复杂逻辑时，不确定性增高，需要深度探索，由easy模式转换为hard模式；在确认求解方向后，模型不确定性降低，不再需要深度思考，切换回easy模式。

In pilot training experiments, we found that $\mathcal{R}_{\text{mode}}$, which discourages deep reasoning, compresses the exploration space and undermines response accuracy. To analyze the model’s exploration behavior, we measure the entropy at the beginning and end of each reasoning unit. As demonstrated in Figure \ref{fig:ourmethod} (b), the terminal entropy values of easy units are generally higher than their initial entropy values, whereas hard units exhibit the opposite trend, which indicates that transitions from easy to hard mode exhibit higher entropy, consistent with the requirement for deeper exploration. 

% 基于这一现象，我们引入了熵值引导的动态rollout策略：在由easy切换为hard时，生成多个分支以扩展模型的探索空间，通过拓宽探索宽度来弥补探索深度的减少。具体而言，我们记录模型生成第一个hard单元时的初始熵值$H_0$，随后当模型由easy转入hard单元时，根据此时模型的熵值与初始熵值的差值$\Delta H$，以概率$\alpha + \Delta H$决定模型是否在此处产生分支，如\ref{fig:ourmethod}所示。
Based on this observation, we propose an \textbf{E}ntropy-guided \textbf{D}ynamic \textbf{R}ollout (\textbf{EDR}) strategy: when transitioning from easy to hard mode, model generates multiple branches to expand its exploration space, compensating for reduced reasoning depth by increasing reasoning breadth.Specifically, we record the entropy of the first k tokens as the initial entropy $H_0$ when generating the first hard unit. Upon transitioning from an easy to a hard unit, the model branches with probability $\alpha + \Delta H$, where $\alpha=0.5$ is the base probability and $\Delta H$ is the normalized entropy difference, as illustrated in \ref{fig:ourmethod}(c). 

\section{Experiments}

% \footnotetext[1]{\url{https://huggingface.co/datasets/yentinglin/aime_2025}} 
% \footnotetext[2]{\url{https://huggingface.co/datasets/HuggingFaceH4/aime_2024}}
% \footnotetext[3]{\url{https://huggingface.co/datasets/zwhe99/amc23}}

\subsection{Experimental Setup}

We build the cold-start dataset with 300k examples sampled from the OpenMathReasoning \citep{moshkov2025aimo2} dataset, then conduct our EHPO training on the DeepScaleR-Preview \citep{deepscaler2025} dataset and adopt a two-stage training procedure with max response length limits of 8k and 16k tokens following DeepScaleR. Note that the 8k stage is intended to rapidly strengthen the model's foundational capabilities, and thus the entropy-guided dynamic rollout strategy is applied exclusively in the 16k stage. For evaluation, we use four mathematical reasoning benchmarks: AIME25~\cite{AIME2025}, AIME24~\cite{AIME2024}, and MATH500~\cite{MATH500}.

We use DeepSeek-R1-Distill-Qwen-1.5B as the base model and compare against the following baselines: (1) O1-Pruner \citep{luo2025o1}, a fine-tuning method that uses pre-sampling and RL-style optimization to reduce reasoning length while preserving accuracy in LRMs; (2) DRP \citep{DRP}, a distillation–pruning framework that reduces token usage via teacher-guided step pruning; (3) Efficiency Steering (ES) \citep{EfficiencySteering}, leveraging large models’ intrinsic potential to produce concise reasoning while preserving accuracy;(4) ACPO \citep{ACPO} also trains models to switch reasoning modes; but in contrast to our approach, it adopts standard GRPO with a customized reward function, without enforcing a strict distinction between the two modes during RL training.

\subsection{Results}

\paragraph{Balancing Reasoning Efficiency and Accuracy} As shown in Table~\ref{sample-table}, our method achieves strong performance across datasets. On challenging tasks, it attains the highest accuracy on AIME2024 (36.5\%, 6.1\% higher than baseline) with 50.3\% shorter outputs, yielding the best efficiency score of 1.10, and maintains competitive accuracy on AIME2025 with 49.5\% fewer tokens with AES of 0.45. On MATH500, it preserves accuracy while reducing token usage by nearly 60\% with AES of 0.55. Overall, our approach achieves the best average AES of 0.70. Note that as DRP and ACPO did not provide results on the most challenging AIME2025, their average AES is likely upward-biased, whereas ADR still outperforms the strongest baseline DRP (0.68).

\paragraph{Ablation of Entropy-Guided Dynamic Rollout Strategy} Unoptimized RL training (ADR w/o EDR) brings only limited benefits, reaching 33.8\% accuracy on AIME2024 with an average AES of just 0.51. In contrast, adding EDR significantly improves both accuracy and efficiency: on AIME2024, accuracy rises to 36.5\% (2.7\% higher than ADR w/o EDR) with the highest AES of 1.10, and similar efficiency gains are observed across tasks. Overall, EDR boosts the Avg. AES from 0.51 to 0.70, confirming that EDR enables more effective accuracy–efficiency trade-offs.

\begin{table}
\setlength{\tabcolsep}{2pt}
    \caption{Performance comparison of various baselines and our method. The \textbf{bold} and \underline{underlined} values denote the best and second-best results, respectively. Accuracy-Efficiency Score (AES), introduced by O1-Pruner \citep{luo2025o1}, measures efficiency by rewarding shorter outputs without compromising accuracy. The Avg AES is computed over benchmarks with reported results, excluding tasks where outcomes are unavailable. The accuracy (Acc.) is measured by the pass@1, which is estimated as the average correctness over 16 sampled generations.}
    \label{sample-table}
    \centering
    \begin{tabular}{l ccc ccc ccc c}
        \toprule
         & \multicolumn{3}{c}{\bf AIME2025} & \multicolumn{3}{c}{\bf AIME2024} & \multicolumn{3}{c}{\bf MATH500} \\
        \cmidrule(lr){2-4} \cmidrule(lr){5-7} \cmidrule(lr){8-10} 
        & \bf Acc. & \bf Tokens & \bf AES & \bf Acc. & \bf Tokens & \bf AES & \bf Acc. & \bf Tokens & \bf AES & \bf Avg. AES \\
        \midrule
        \bf Baseline & 23.5 & 12119 & --- & 30.4  & 12290 & --- & 81.7  & 4802 & --- & --- \\
        \midrule
        \bf O1-Pruner & 23.2  & 8731\textsubscript{$\downarrow$28.0\%} & 0.22 & --- & --- & --- & \textbf{84.3} & 2913\textsubscript{$\downarrow$39.3\%} & 0.49 & 0.35 \\
        \bf DRP & --- & --- & --- & 33.3  & 6135\textsubscript{$\downarrow$50.1\%} & 0.79 & 82.0  & 2122\textsubscript{$\downarrow$55.8\%} & 0.57 & \underline{0.68} \\
        \bf ES & \textbf{24.2}  & 7458\textsubscript{$\downarrow$38.5\%} & \textbf{0.47} & 28.3  & 9024\textsubscript{$\downarrow$26.6\%} & -0.08 & \underline{83.0}  & 2400\textsubscript{$\downarrow$50.0\%} & 0.55 & 0.31 \\
        \bf ACPO & --- & --- & ---& 30.0  & 6670\textsubscript{$\downarrow$45.7\%} & 0.39 & 81.0 & \textbf{1679}\textsubscript{$\downarrow$65.0\%} & \textbf{0.61} & 0.50\\
        \midrule
        \bf ADR w/o EDR & 21.5  & \textbf{5890}\textsubscript{$\downarrow$51.4\%} & 0.09 & \underline{33.8}  & \textbf{5971}\textsubscript{$\downarrow$51.4\%} & \underline{0.85} & 81.6 & 1992\textsubscript{$\downarrow$58.5\%} & \underline{0.58} & 0.51\\
        \bf ADR & \underline{23.3}  & \underline{6126}\textsubscript{$\downarrow$49.5\%} & \underline{0.45} & \textbf{36.5}  & \underline{6110}\textsubscript{$\downarrow$50.3\%} & \textbf{1.10} & 81.0  & \underline{1955}\textsubscript{$\downarrow$59.3\%} & 0.55 & \textbf{0.70} \\
        \bottomrule
    \end{tabular}
\end{table}

% \begin{table}
%     \caption{Ablation of Entropy-Guided Dynamic Rollout Strategy}
%     \label{sample-table}
%     \centering
%     \begin{tabular}{l cc cc cc cc}
%         \toprule
%         & \multicolumn{2}{c}{aime2025} & \multicolumn{2}{c}{aime2024} & \multicolumn{2}{c}{math500} & \multicolumn{2}{c}{AMC23} \\
%         \cmidrule(lr){2-3} \cmidrule(lr){4-5} \cmidrule(lr){6-7} \cmidrule(lr){8-9}
%         & acc & len & acc & len & acc & len & acc & len \\
%         \midrule
%         Base model & \textbf{23.5} & 12119 & 30.4  & 12290 & 81.7  & 4802 & 59.6 & 8620 \\
%         % +SFT & 19.6 & 11553 & 27.7 & 11715 & \textbf{82.0} & 3778 & 61.7 & 7338 \\
%         +RL w/o EDR & 21.5  & \textbf{5890} & 33.8  & \textbf{5971} & 81.6 & 1992 & 65.2 & \textbf{3790} \\
%         +RL w/ EDR & 23.3  & 6126 & \textbf{36.5}  & 6110 & 81.0  & \textbf{1955} & \textbf{65.7} & 3790 \\
%         \bottomrule
%     \end{tabular}
% \end{table}

\subsection{Conclusions}

In this work, we present \textbf{ADR}, a new reasoning paradigm for large reasoning models, enabling model dynamically switches between fast reasoning for straightforward cases and intensive reasoning for complex dependencies. To optimize reasoning allocation, we introduce \textbf{EHPO}, which combines mode control reward with entropy-guided dynamic rollout to expand the exploration space while maintaining accuracy. Extensive experiments on multiple mathematical reasoning benchmarks show that our approach achieves an effective balance between reasoning efficiency and accuracy, demonstrating robust performance across tasks of varying difficulty.

{
\small

\setcitestyle{numbers,square}
\bibliographystyle{unsrt}
\bibliography{reference}

% [1] Alexander, J.A.\ \& Mozer, M.C.\ (1995) Template-based algorithms for
% connectionist rule extraction. In G.\ Tesauro, D.S.\ Touretzky and T.K.\ Leen
% (eds.), {\it Advances in Neural Information Processing Systems 7},
% pp.\ 609--616. Cambridge, MA: MIT Press.

% [2] Bower, J.M.\ \& Beeman, D.\ (1995) {\it The Book of GENESIS: Exploring
%   Realistic Neural Models with the GEneral NEural SImulation System.}  New York:
% TELOS/Springer--Verlag.

% [3] Hasselmo, M.E., Schnell, E.\ \& Barkai, E.\ (1995) Dynamics of learning and
% recall at excitatory recurrent synapses and cholinergic modulation in rat
% hippocampal region CA3. {\it Journal of Neuroscience} {\bf 15}(7):5249-5262.
}

%%%%%%%%%%%%%%%%%%%%%%%%%%%%%%%%%%%%%%%%%%%%%%%%%%%%%%%%%%%%

\appendix

% \section{Technical Appendices and Supplementary Material}
% Technical appendices with additional results, figures, graphs and proofs may be submitted with the paper submission before the full submission deadline (see above), or as a separate PDF in the ZIP file below before the supplementary material deadline. There is no page limit for the technical appendices.

\section{Details of Cold-Start Data Construction}

\subsection{Datasets}

We construct the cold-start dataset based on the open-source mathematical reasoning corpus \citep{moshkov2025aimo2}, which comprises 306k math problems and 3.2 million long CoT responses generated by Deepseek-R1 and QwQ-32B. From this corpus, we extract and reformulate 300k examples for cold-start SFT training. 

\subsection{Prompts}

\paragraph{Prompt for ADR Generating} We provide the prompt we use for ADR-paradigm reasoning:

\begin{figure}[htbp]
    \centering
    \includegraphics[width=0.9\textwidth]{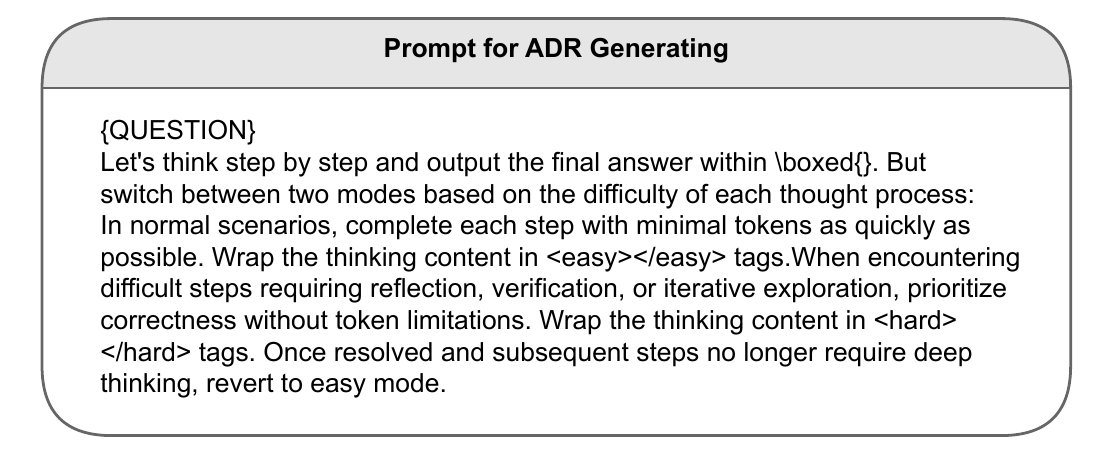}
    \caption{Prompt for ADR Generating}
    \label{fig:ADR prompt}
\end{figure}

\begin{figure}[htbp]
    \centering
    \includegraphics[width=0.9\textwidth]{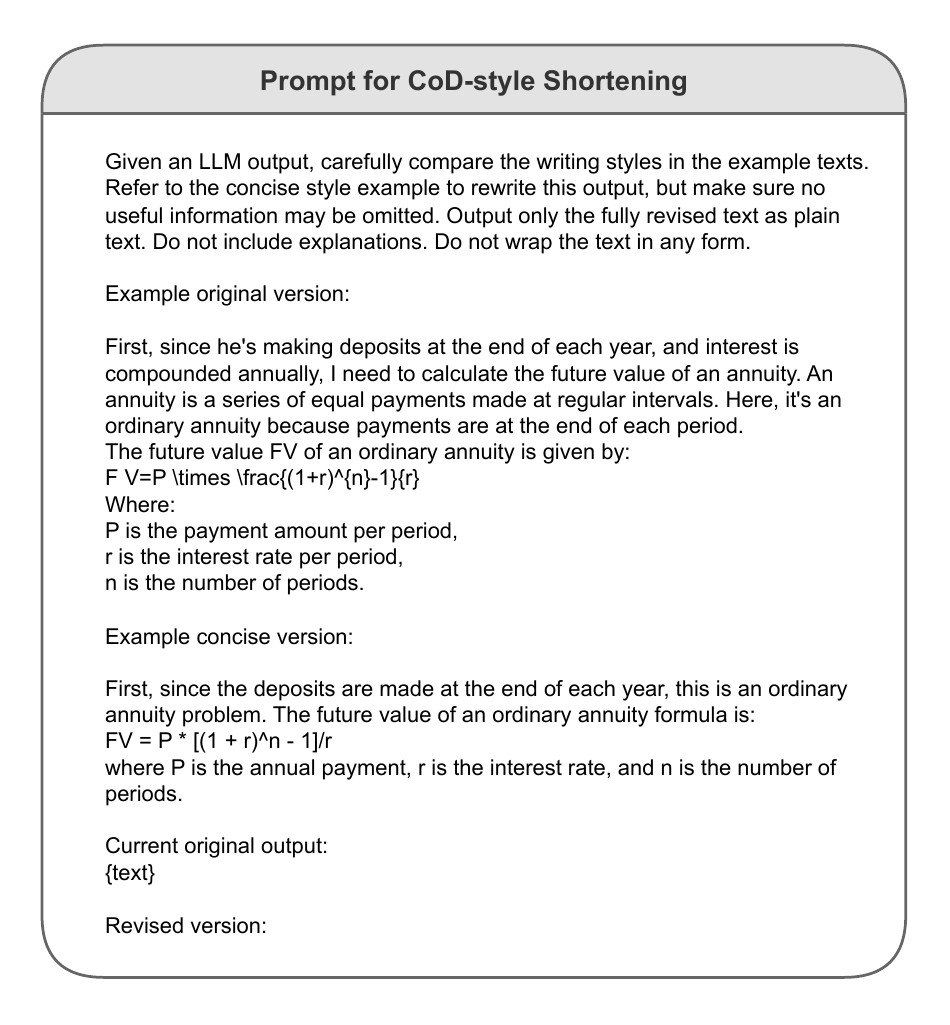}
    \caption{Prompt for CoD-Style Shortening}
    \label{fig:shortening prompt}
\end{figure}

\paragraph{Prompt for CoD-Style Shortening} We use Deepseek-R1-0528 to shorten the content of easy units in CoD-style \citep{CoD}. As shown in \ref{fig:shortening prompt}, we provide a paired comparison example between a real CoD-style output and the original R1-style output as a reference.

\section{Case Study}

\begin{figure}[htbp]
    \centering
    \includegraphics[width=0.75\textwidth]{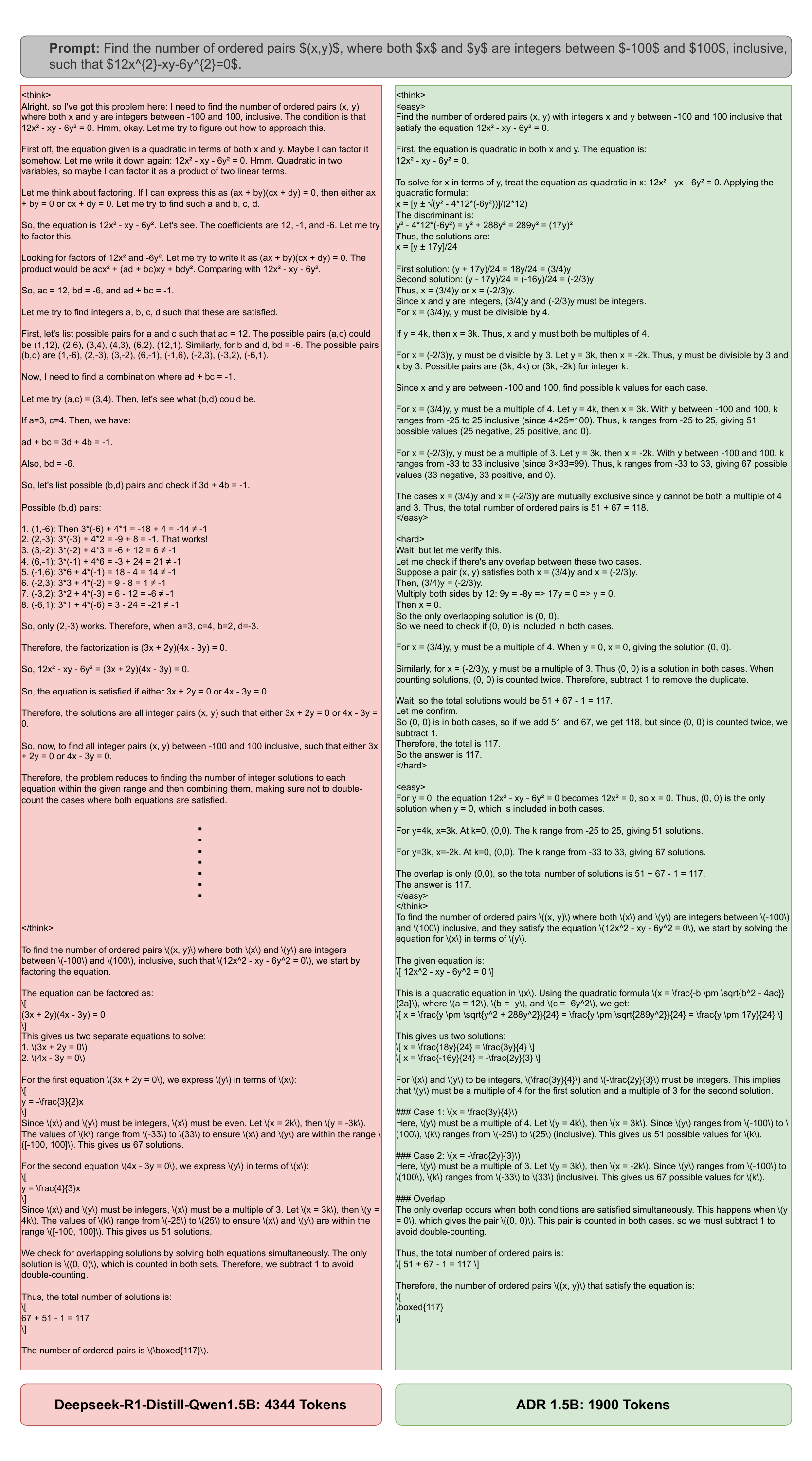}
    \caption{A case study comparing the reasoning process of DeepSeek-R1-Distill-Qwen-1.5B and ADR in AIME2025.}
    \label{fig:case}
\end{figure}

As illustrated in Figure \ref{fig:case}, we observe a striking contrast between the baseline model and our ADR-trained model. The baseline (Deepseek-R1-Distill-Qwen1.5B) produced an output of 4344 tokens, reflecting an almost exclusive reliance on the deep reasoning. Its reasoning process was verbose and exploratory: the model repeatedly attempted factorization, re-derived discriminants, and engaged in extensive verification, even when the solution path had already been established. While correct, this exhaustive style incurred substantial inefficiency.

By contrast, the ADR-trained model (ADR 1.5B) completed the same task in only 1900 tokens. The improvement stems from ADR’s ability to dynamically switch between fast and slow thinking. Most steps were carried out in the easy mode—directly computing discriminants, deriving closed-form solutions, and quickly mapping integer constraints, exhibiting higher information density. Compared with the baseline, ADR reaches the same intermediate conclusions using fewer tokens. The model switched into hard mode only at crucial junctures, such as verifying overlaps between solution families in this case. This selective use of deeper reasoning preserved correctness while avoiding unnecessary elaboration.

Overall, ADR does not merely compress outputs; rather, it strategically allocates slow thinking only where needed, yielding concise yet rigorous reasoning. This case study highlights ADR’s effectiveness in substantially improving inference efficiency without sacrificing accuracy.

%%%%%%%%%%%%%%%%%%%%%%%%%%%%%%%%%%%%%%%%%%%%%%%%%%%%%%%%%%%%

\end{document}